\documentclass[journal,onecolumn]{style/IEEEtran}

\usepackage{lineno}
\usepackage{soul}
\usepackage{xspace}
\usepackage[normalem]{ulem}
\usepackage{color}

\usepackage{amsmath}
\usepackage{amssymb}
\usepackage{amstext}
\usepackage{amsfonts}
\usepackage{algorithm}
\usepackage[noend]{algorithmic}
\usepackage{graphicx}
\usepackage{multirow}
\usepackage{tabularx}
\usepackage{tensor}
\usepackage{rotating}
\usepackage{url}
\usepackage{ulem}

\graphicspath{{images/}}

\newcommand{\superscript}[1]{\ensuremath{^\textrm{#1}}}
\newcommand{\subscript}[1]{\ensuremath{_\textrm{#1}}}

\newcommand{\etal}{\emph{et al.}\mbox{ }}
\renewcommand{\paragraph}[1]{\vspace{0.5em} \noindent \textit{#1:}}

\def\cf{\textit{c.f.}}
\def\etal{\textit{et al.}\xspace}

\renewcommand{\emph}[1]{\textit{#1}}

\begin{document}
\title{Towards MRI-Based Autonomous Robotic US Acquisitions: A First Feasibility Study}
\author{Christoph~Hennersperger*,
	Bernhard~Fuerst*,
	Salvatore~Virga*,
	Oliver~Zettinig,
	Benjamin~Frisch,
	Thomas~Neff,
        and~Nassir~Navab
\thanks{*Authors contributed equally. Corresponding author e-mail: christoph.hennnersperger@tum.de | 
C. Hennersperger, S. Virga, O. Zettinig, B. Frisch, and N. Navab are
 with the Chair for Computer Aided Medical Procedures, Technische Universit\"at M\"unchen, Germany. T. Neff and S. Virga are further with KUKA Roboter GmbH, Augsburg, Germany, and B. Fuerst and N. Navab are further with Computer Aided Medical Procedures, Johns Hopkins University, Baltimore, USA.
}
}

\markboth{
}{
Hennersperger, Fuerst, Virga, Navab: Autonomous Robotic US Acquisitions
}

\maketitle

\begin{abstract}
\noindent Robotic ultrasound has the potential to assist and guide physicians during interventions. In this work, we present a set of methods and a workflow to enable autonomous MRI-guided ultrasound acquisitions. Our approach uses a structured-light 3D scanner for patient-to-robot and image-to-patient calibration, which in turn is used to plan 3D ultrasound trajectories. These MRI-based trajectories are followed autonomously by the robot and are further refined online using automatic MRI/US registration. Despite the low spatial resolution of structured light scanners, the initial planned acquisition path can be followed with an accuracy of $2.46 \pm 0.96$~mm. This leads to a good initialization of the MRI/US registration: the 3D-scan-based alignment for planning and acquisition shows an accuracy (distance between planned ultrasound and MRI) of $4.47$~mm, and $0.97$~mm after an online-update of the calibration based on a closed loop registration.
\end{abstract}


 \ifCLASSOPTIONpeerreview
 \begin{center} \bfseries EDICS Category: 3-BBND \end{center}
 \fi
%
%
%
\section{Introduction}
\label{sec:introduction}
%
%
\IEEEPARstart{U}{ltrasound} (US) has become one of the standard medical imaging techniques and is widely used within diagnostic and interventional applications. 
During needle insertion for liver biopsy or ablation, ultrasound imaging is used today in order to guide the insertion process throughout the procedure. 
Based on pre-interventional X-ray Computed Tompgraphy (CT) or Magnetic Resonance Imaging (MRI) datasets, a needle is advanced carefully under US-guidance to reach a final target position.
Thereby, it is important to maintain steady ultrasound views throughout the intervention, providing an overview of all essential structures in the abdomen (tumor, vessels, lung, liver tissue).

In general, clinical US is mostly based on 2D-images (except cardiac and gynaecological applications), requiring a manual navigation of the probe.
The resulting high operator-variabilty of the manual guidance is not only challenging for the application described above, but impairs a wider clinical acceptance of ultrasound for the extraction of quantifiable parameters from these data~\cite{beales2011reproducibility}.
3D ultrasound can potentially overcome these limitations, and is performed either using native 3D probes or by tracking 2D images in space, using a tracking target attached to the ultrasound probe (tracked freehand 3D ultrasound)~\cite{gee2003engineering}.
While systems using native 3D probes are still not widely available in clinical practice, tracked ultrasound is easily accessible and can also be interpolated with respect to a regular grid~\cite{solberg2007freehand}.
When comparing native and freehand 3D ultrasound, both techniques have their merits. 
Native 3D ultrasound allows, on the one hand, for live 3D volume acquisitions in real-time and thus for a direct analysis of volume changes over time; a property which is especially exploited for 3D echocardiography~\cite{monaghan2006role}. 
On the other hand, the systems are expensive and only allow for the imaging of comparably small volumes restricted by the probes' field of view. 
Freehand 3D ultrasound does not pose limitations with respect to volume sizes, anatomies and trajectories, but can be potentially distorted by varying pressure applied by the operator, or changing anatomy caused by breathing or cardiac pulsation.
\begin{figure}
	\centering
	\includegraphics[width=0.5\columnwidth]{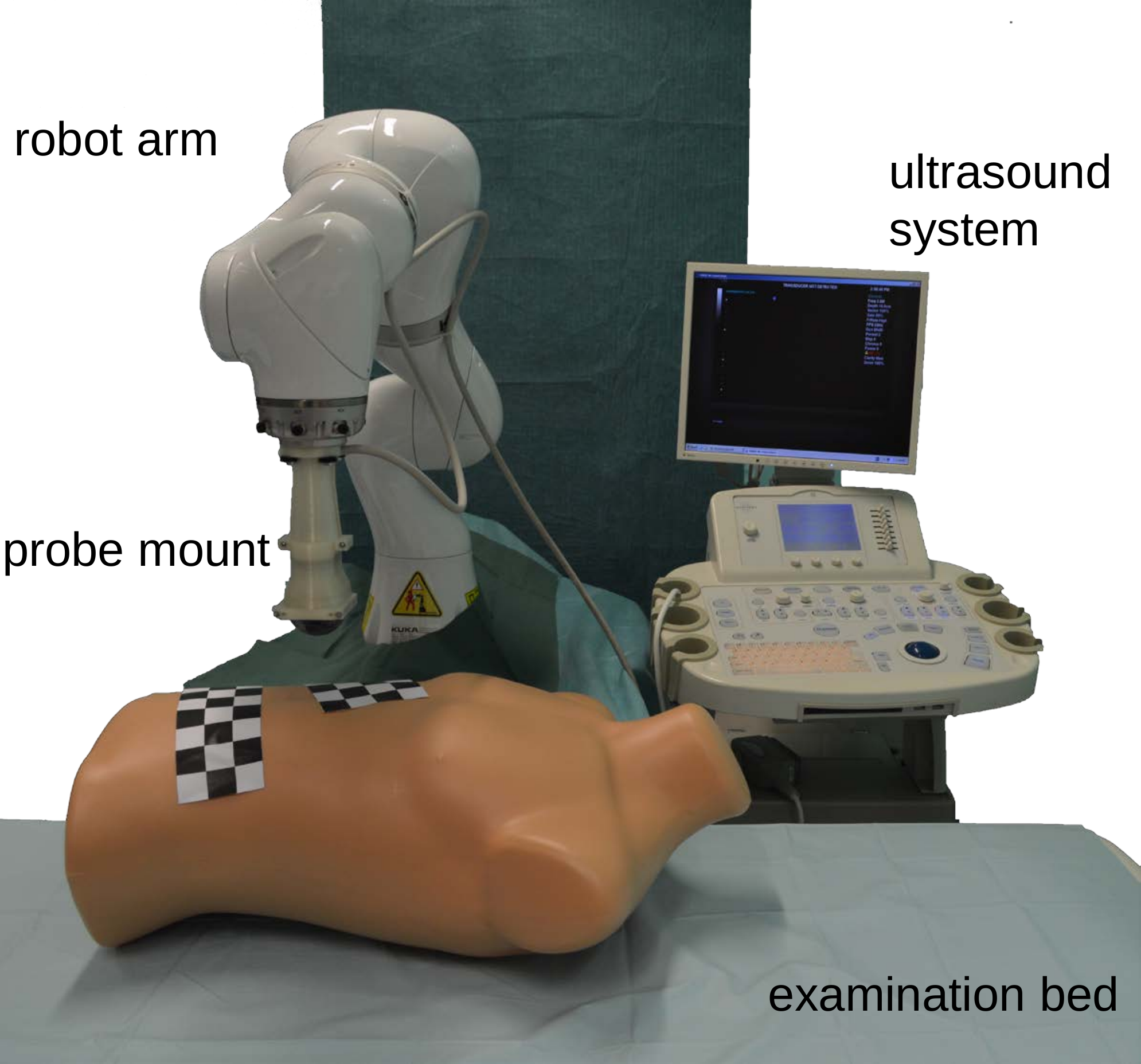}
	\caption{In the presented system setup, a RGBD-camera is mounted on the ceiling, observing the scene with the patient lying on the examination bed. The robotic arm is approaching the patient from above and is equipped with a curvilinear ultrasound probe.}
\end{figure}

While ultrasound acquisitions are mostly performed by physicians or medical staff, data collection for both 2D and 3D ultrasound might be improved by automating the acquisition process.
This goes in hand with the demands of the healthcare system, showing that current freehand ultrasound scanning techniques suffer from high operator variability, hindering the overall clinical acceptance, especially of 3D applications~\cite{beales2011reproducibility,cardinal2000analysis}.

To this end, a fully automatic system to acquire 3D ultrasound datasets could avoid the high inter-operator variability and resource demands of current clinical systems.
In view of liver ablation for example, such automatic ultrasound acquisitons could be performed periodically throughout the procedure to guide the insertion, while CT or MRI datasets will be available for an interventinal planning.
By addressing limitations of today's ultrasound techniques using a robotic framework, this would not only open up the way for new applications in an interventional setting, but also for potential society-wide screening programs using non-invasive ultrasound imaging~\cite{beales2011reproducibility,de2012aneurysm}.
In this view, autonomous acquisitions have the potential both to facilitate clinical acceptance of robotic imaging techniques by simplifying the workflow as well as to reduce the scanning time and the necessary amount of manual operator input.

First attempts aiming at automatized ultrasound acquisitions  used mechanical stepper motors, moving the ultrasound probe in a controlled fashion~\cite{fenster2001three}.
More recently, different robotic systems were developed in the context of clinical ultrasound, including applications to imaging, surgical interventions and needle guidance~\cite{priester2013robotic,chatelain20153d}. 
While freehand ultrasound scanning enables a fast and dynamic acquisition and screening of several anatomies, modern compact and lightweight robotic arms can further support the physician, i.e. by automatically maneuvering a second imaging probe or tool in order to enable live image fusion~\cite{esposito2015cooperative}.
Moreover, such systems eventually incorporate pressure, hand tremor or movement correction~\cite{krupa2009real}, or an automatic servoing for tissue biopsy based on a registration to prior US acquisitions~\cite{zettinig2015a}. 
With the goal of fully automatic acquisitions, however, a prior planning of the actual target area of interest is a prerequisite, since whole-body ultrasound scans are impractical and time-consuming.
In a clinical interventional setup, planning should be performed by the medical expert based on anatomical data, as given by MRI and CT.
Given an appropriate target, (several) 3D ultrasound datasets can then be acquired autonomously during an intervention, without requiring the presence of medical staff guiding the robot.
We acknowledge that there have been attempts in the past to incorporate tomographic image information into robotic systems to improve the visualization of ultrasound image information to physicians.
However, to our knowledge an integration of these data to enable planning and the \emph{automatic} acquisition of 3D US datasets has not been considered so far.

\begin{figure}[t]
	\centering
	\includegraphics[width=0.4\columnwidth]{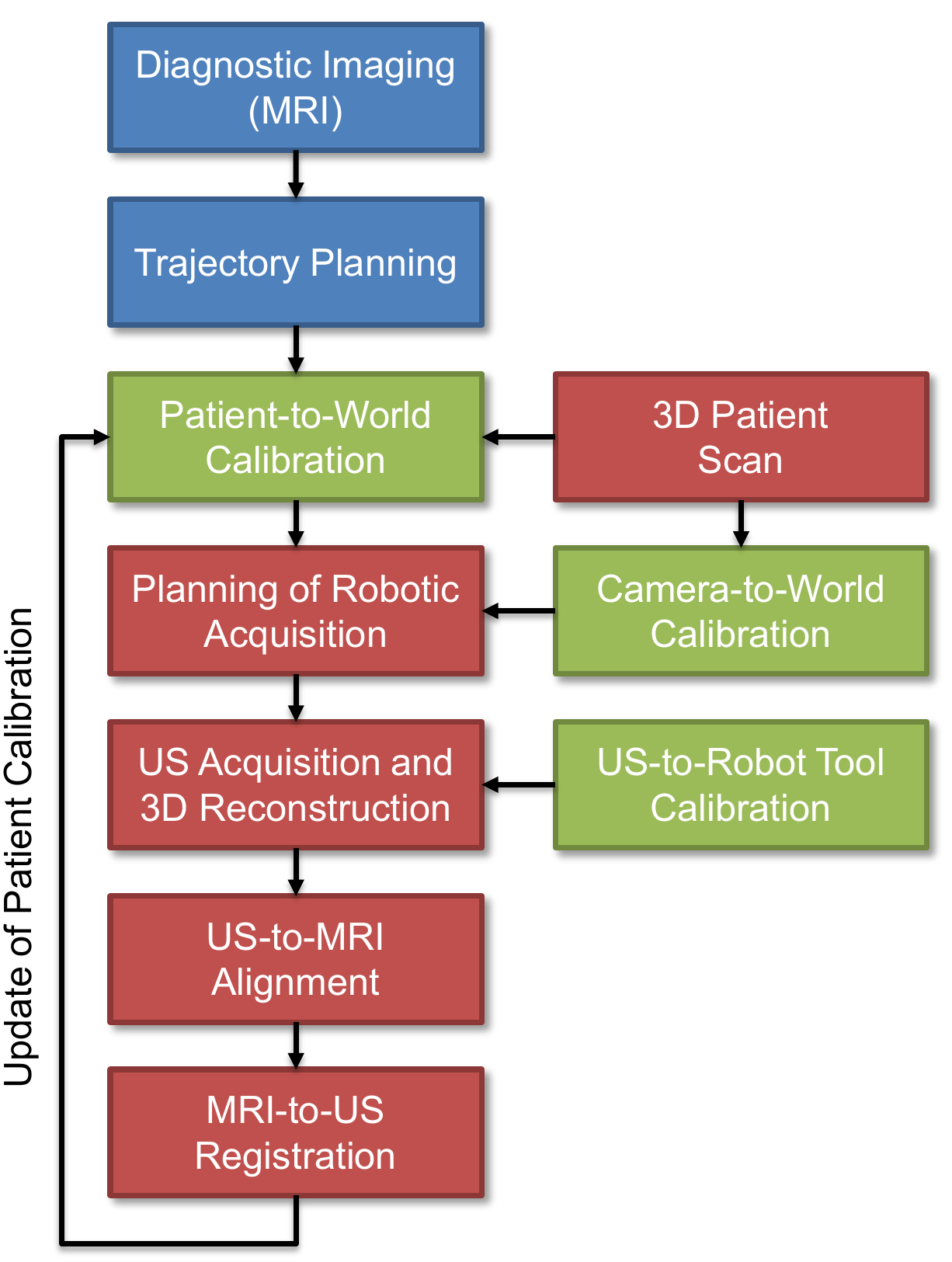}
	\caption{The proposed workflow for autonomous acquisitions includes a planning based on diagnostic images (blue), calibration of both robot and patient to the world reference (green), and interventional acquisition (red). During acquisition, a 3D patient scan using a structured-light 3D scanner is used for the initial patient to world alignment and robot to world calibration. Following the autonomous robotic ultrasound acquisition, a US to MR registration is conducted to refine the patient to world alignment.}
	\label{workflow}
\end{figure}

In this work, we aim at closing this gap in the workflow of current robotic support systems in order to allow for fully automatic 3D ultrasound acquisitions using a robotic imaging system. 
We present the path towards an autonomous robotic ultrasound system to assist clinicians during interventions by performing multiple and especially reproducible examinations based on pre-interventional planning. 
To this end, we introduce a first concept for a robotic ultrasound system consisting of a lightweight robotic arm, a clinical ultrasound machine, and a RGB-D camera which is used to observe the scene.
This enables the direct planning of a patient-specific trajectory by selecting its start- and endpoints in the patient's MR (or CT) image.
A registration of the actual patient position to the MRI  allows for the automatic acquisition of 3D ultrasound data. 
By using intensity-based image registration, we can close the loop and perform an online-update of the patient-to-world registration, accounting for inaccuracies of the RGB-D information as well as patient movement.
Thus, the overall goal is to perform fully autonomous image acquisitions within a closed control loop by utilizing 3D surface information, pressure estimations of the robotic system, and image-based servoing, to image regions of interest defined by pre-interventional data\footnote{Supplementary video material to this manuscript features a brief overview of the acquisition setup and workflow with the distinct steps.}.
The workflow showing the main steps of the proposed solution is depicted in Fig.~\ref{workflow}. 

The remainder of this work is organized as follows: Section \ref{sec:relatedWork} will give an overview of related work in the field of autonomous robotic ultrasound acquisitions. We introduce our proposed system and all corresponding and necessary steps in Section \ref{sec:methods}, before experiments and results are presented in Section \ref{sec:experiments}. Finally, opportunities and challenges associated to the system are discussed in Section \ref{sec:discussion}, before we draw final conclusions in Section \ref{sec:conclusion}.

%
%
\section{Related Work}
\label{sec:relatedWork}
%
%
\noindent Over the past two decades, research groups have focused on improving ultrasound image acquisition systems in terms of accuracy, usability, remote controllability as well as by a synchronization with pre-operative planning.
For this purpose, (medical) robots have been introduced for different applications. Inspired by~\cite{priester2013robotic}, we differentiate between situation aware robotic systems, and surgical assistant systems.

Surgical assistant systems primarily serve as advanced tools and are directly controlled by the surgeon, mostly acting in a master-slave fashion. 
Thereby, these systems are used to augment the operator's ability to perform a specific task, e.g. by allowing precise, jitter-free movements, or the application of constant desired pressure.
Systems targeted to specific applications, such as carotid imaging~\cite{nakadate2011development}, have been explored recently. 
Furthermore, ultrasound has been added as a tool to the more general surgical da Vinci platform for several applications~\cite{billings2012system,schneider2011intra} using specialized drop-in ultrasound probes which can be picked up by the surgeon intra-operatively. 
Following this concept, a general system providing tele-operated ultrasound using different robotic arms is currently being developed~\cite{conti2014interface}.
For a more detailed overview of surgical assistants in ultrasound, the reader is referred to~\cite{priester2013robotic}.

In contrast to these systems, situation aware robotic system perform at least one task autonomously based on a predefined situation-specific model, requiring an awareness of the task and its surrounding environment.
In the following, we will focus our review on this area, as it is the primary target area of this work. 
In order to do so, we differentiate between Automatic Robotic Support Systems (ARSS) providing automatic support for a defined task, as well as Automatic Data Acquisition Systems (ADAS), targeting fully automatic robotic imaging.

\subsection{Automatic Robotic Support Systems}
Boctor~\etal~\cite{boctor2004dual} originally proposed a dual robotic arm setup holding a needle in one arm and a US probe in the second, in order to facilitate an accurate placement of the needle to perform liver ablations. 
In a later work~\cite{boctor2008three}, the robotic arm holding the US probe was replaced by a freehand 3D ultrasound setup to improve general usability while still enabling accurate placement.
With respect to an integration of tomographic image data into the robotic ultrasound environment, Zhang~\etal~\cite{zhang2013optical} propose a system combining a robotic arm with ultrasound, optical tracking and MRI data manually registered to the system in order to improve the accuracy of needle placement.
Targeting at optimal robotic ultrasound acquisitions to cover a predefined volume of interest, a planning-framework is presented in \cite{hennersperger2016a} to perform automatic robotic imaging trajectories.
Other groups have focused more on a direct integration of the resulting images into the robot control, facilitating visual servoing based on robotic ultrasound. Abolmaesumi~\etal~\cite{abolmaesumi2000visual} were among the first exploring the combination of robotic arms and ultrasound imaging based on visual servoing by combining an experimental robotic arm holding an ultrasound probe with vessel feature tracking to follow vessels during the ultrasound scan.
Krupa~\etal~\cite{krupa2009real} further explored visual servoing towards an application to ultrasound by using ultrasonic speckle as a main characteristic feature to compensate for potential patient motion both in-plane and out-of-plane with respect to the ultrasound probe. 
These concepts were later adapted to fully automatic needle insertion under ultrasound guidance~\cite{krupa20143d} and automatic probe positioning in order to allow for an optimal skin coupling with respect to the current position~\cite{chatelain2015optimization}.
Recently, a servoing approach using live ultrasound to ultrasound registration was presented, targeting at screw placement in spine surgery~\cite{zettinig2015a, zettinig2015b}.
Another intra-operative application of robotic ultrasound was presented for the mobilization and the resection of the internal mammary artery in the course of coronary artery bypass grafting~\cite{frohlich2010robot}, following a servoing approach using color Doppler ultrasound images.
For a comprehensive and detailed background of visual servoing, the reader is further referred to~\cite{chaumette2006visual,chaumette2007visual}, where concepts and basic approaches are explained in detail.

\subsection{Automatic Data Acquisition Systems}
While all approaches described above perform ultrasound acquisitions semi-automatically or under direct physician guidance, they either require a manual positioning of the US probe, or a proper definition of the working area. 
A first step towards fully autonomous scanning of ultrasound trajectories~\cite{onogi2013robotic} combines a designed pneumatic probe holding case with optical tracking in order to enable scanning of a small volume of interest. 
While this approach has a high potential for certain applications, it still requires manual placement of the autonomous holding cage on the patient and does not enable fully automatic scanning.
Using two robotic arms holding separate ultrasound probes, a system for 3D robotic tomographic imaging pioneers the idea of localized 3D tomographic ultrasound imaging~\cite{aalamifar2014enabling}.
With respect to an intra-operative setting, however, the positioning of two robotic arm seems challenging, which is why the major field of applications for this technique might lie in the imaging of smaller structures.
Focusing on vascular applications, a targeted scanning system for the automatic acquisition of vascular 3D ultrasound is presented in \cite{merouche2016robotic}, combining pressure compensation with vessel tracking mechanisms.
Finally, another system focusing on automatic data acquisition for liver screening was recently presented~\cite{mustafa2013development}, combining RGB cameras for an automatic determination of liver scanning regions, followed by an automated acquisition workflow using different scanning directions. While this system operates fully automatically, it is limited to 2D images only and performs acquisition based on classified body features, which are prone to an erroneous feature determination, limiting the practical applicability of the system.

With this work, we focus explicitly on a system which enables fully autonomous robotic 3D ultrasound acquisitions based on an individual planning from tomographic image data (performed before the intervention).
Thereby, the system does not rely on an initial probe positioning, manual registration of image data, or feature tracking in ultrasound or RGB-images. 
Instead, we use an initial patient-to-robot registration to perform 3D US acquisitions, which are then used in order to refine the overall registration and calibration.
To do so, an intensity-based image registration of the pre-aligned 3D US and tomographic image data is used to retrieve a refined and updated patient-to-robot registration.
Using the robot's built-in force and torque sensors, this further allows ultrasound acquisitions with optimal skin force, providing repeatable 3D ultrasound volumetric datasets.
%
%
%
%
\section{Methods}\label{sec:methods}
%
%
\noindent This section first introduces the main component of the robotic ultrasound hardware setup in \ref{sec:system}. Next, \ref{sec:mri} describes the pre-interventional imaging as well as trajectory planning for autonomous acquisitions. 
Following our proposed application, all required calibration steps and procedures are defined in \ref{sec:calibration}, before the interventional workflow for robotic acquisitions is explained in detail in \ref{sec:acquisition}, including robotic control strategies, updates, ultrasound acquisitions and refinement of world-calibrations based on US acquisitions.

\subsection{Hardware Setup}\label{sec:system}
Our system consists of a lightweight robot, the ultrasound device, and a structured light RGB-D 3D scanner. While the ultrasound transducer is directly mounted onto the end-effector of the robotic arm, the 3D scanner is attached to the ceiling and serves as a vision system allowing for the direct calibration and registration of all system parts.
\\

\subsubsection{Robotic Arm}
Based on developments of the German space center (DLR)~\cite{hirzinger2002dlr}, KUKA introduced a robotic platform targeted at direct human-machine interaction, referred to as 'Intelligent industrial work assistant' - iiwa (KUKA Roboter GmbH, Augsburg, Germany).
This system consists of a 7 joint robotic arm with corresponding control units and consequently enables one redundant degree of freedom (6+1 in total). 
As a result of this design, the robot provides dynamic movement and flexible adaption of trajectories to the working environment. 
With respect to robotic ultrasound, the incorporated high-accuracy torque sensors in each of the seven joints are evenly important, as a robotic ultrasound platform has to be fully compliant to both patient and staff. 
Based on significant safety measures for collision detection, the robot subsystem is certified for human-machine-collaboration due to the compliance to standards for functional safety. Thus it is considered to be safe for use in direct interaction with humans.
Detailed specifications and design choices can be found in~\cite{bischoff2010kuka}.

The KUKA native Sunrise.OS and its Sunrise.Connectivity
module allow for the full low-level real-time control of the KUKA iiwa via UDP at rates up to $1$~kHz, acting similar to the Fast Research Interface (FRI) \cite{schreiber2010fast} as proposed for the previous generation of the KUKA LWR robot arms.
In this work, a publicly available software module\footnote{\url{https://github.com/SalvoVirga/iiwa_stack}} developed in our lab is utilized to enable a direct interaction between Sunrise.OS and the Robot Operating System\footnote{\url{http://www.ros.org/}} (ROS) framework.
By doing so, low-level functionality and control provided by the robot manufacturer can be integrated with RGD-D information and high-level robotic interaction through ROS, as required for the proposed system.
\\

\subsubsection{Structured-Light 3D Scanner}
\label{sec:rgbdScanner}
The Kinect 3D camera
(Microsoft Corporation, Redmond, Washington, USA) allows for the recording of color images and depth data (RGB-D) at 30 frames per second. 
The system contains one camera for color video, an infrared laser projector for the projection of a structured light pattern and a camera to record said pattern.
This device has been used for several medical applications, such as for a touch-less interface~\cite{ebert2012you}, for image-to-patient registration~\cite{hsieh2013non}, or to provide initialization to US/MRI registration~\cite{billings2011hybrid}. 
In the proposed system setup, the camera is attached to the ceiling, focusing on the robotic system and examination bed.
The RGB-D data is used in order to i) align the camera within the world coordinate system (camera-to-world calibration - Sec. \ref{sec:robotworldcalib}), and to ii) register the patient lying on the examination bed to the world reference frame (patient-to-world calibration - Sec. \ref{sec:patientworldcalib}). 
\\

\subsubsection{Ultrasound System}
In general, any ultrasound device could be incorporated within an autonomous system by using the video output of an arbitrary clinical US device and recording those images using hardware frame-grabbers connected to the device.
In practice, such systems provide a partially diminished image quality due to compression artifacts and - more importantly for robotic acquisitions - a temporal lag between the acquired image by the US machine and the recorded frame via frame-grabber.
To enable direct acquisitions and advanced servoing in real-time, we favor ultrasound devices providing direct interfaces for retrieving acquired ultrasound data with minimal temporal delay. 
An Ultrasonix RP ultrasound machine (Analogic Corporation, Peabody, Massachusetts, USA) is used in combination with a 4DC7-3/40 4D Convex curvilinear transducer used in single-plane mode. 
The system provides the ulterius\footnote{\url{http://www.ultrasonix.com/wikisonix/index.php/Ulterius}} API, enabling both the streaming of ultrasound data to a client as well as the control of US acquisitions parameters remotely through the Ethernet connection.

\begin{figure}[t]
	\centering
		\includegraphics[width=0.5\columnwidth]{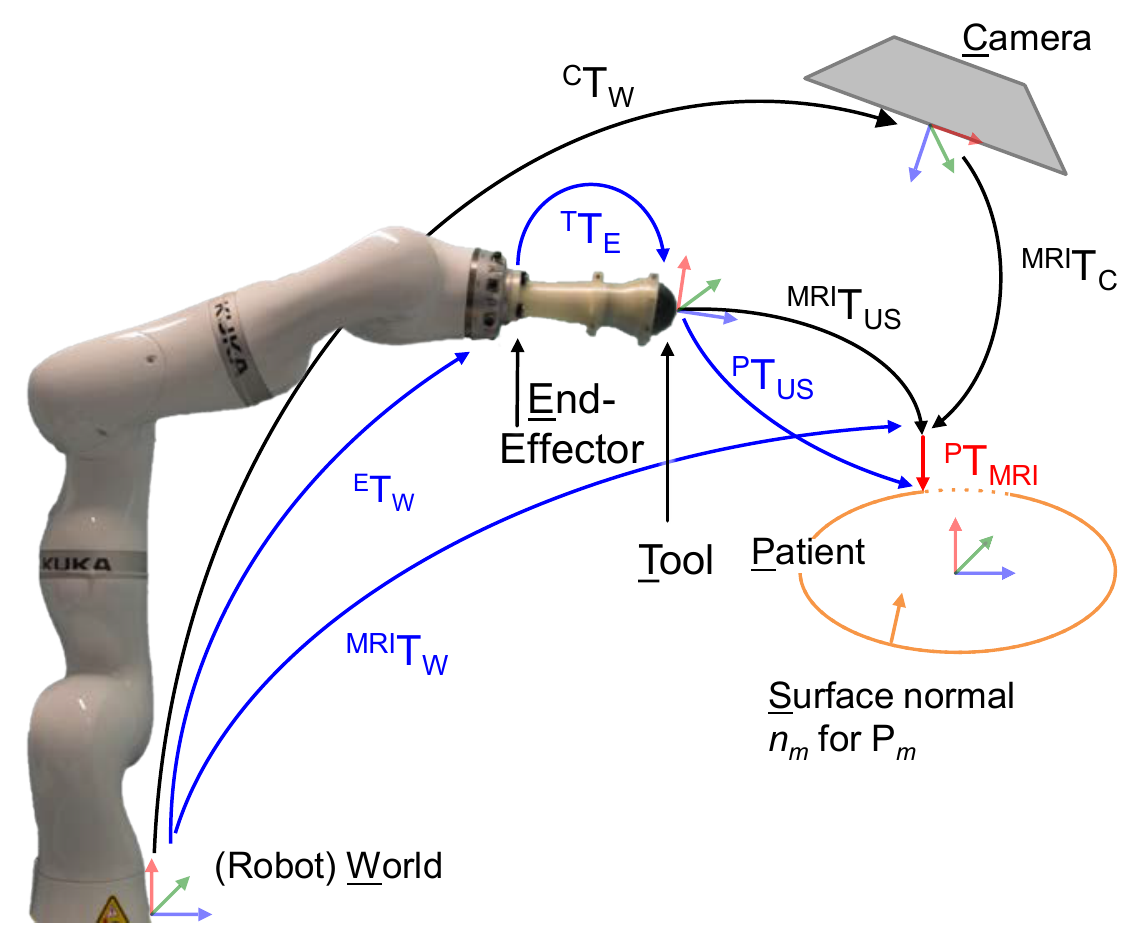}
	\caption{Using the 3D scan of the patient $\tensor[^{MRI}]{\textbf{T}}{_C}$ and the transformation from world to the camera $\tensor[^{C}]{\textbf{T}}{_W}$ the world-to-patient transformation $\tensor[^{MRI}]{\textbf{T}}{_{W}}$ is estimated. The tool (transducer apex) reaches the patient's surface (orange) by applying the tool-to-patient transformation under consideration of the surface normals $n_m$. The discrepancy between the estimated position and the real position of patient is indicated by  $\tensor[^{P}]{\textbf{T}}{_{MRI}}$, which is detected by intensity-based registration (red).}
	\label{setup}
\end{figure}

\subsection{Diagnostic Imaging and Trajectory Planning}\label{sec:mri}

The goal of an autonomous 3D ultrasound acquisition is to perform a robotic-ultrasound trajectory based on a planning on pre-interventional images. 
With respect to a fast and intuitive planning, a physician wants to see either i) the region around a specific target, such as for liver or lymph node biopsy, or ii) scan a defined region, such as an organ, in order to perform a diagnosis based on these images.
In this work we focus on the general case, where the operator can select the region of interest (ROI) directly.
For our setup, a T2-weighted MRI volume is used as basis to determine the ROI for the ultrasound scan. 
The physician simply selects the start- and endpoint $P_s,P_e$ of a trajectory in the MRI data, where the acquisition path direction $\vec{d}_t$ is defined by these points:
\begin{equation}
 \vec{d}_t = P_e - P_s.
 \label{eq:trajectoryDir}
\end{equation}
To enable both the transfer of the planned acquisition path to the robot world coordinate system and the initialization of the MRI/US registration, the patient's surface is extracted from the MRI by thresholding the image. 
Based on the patient-to-world calibration (see Section \ref{sec:patientworldcalib}), the trajectory direction and points can be directly transformed into world coordinates.
Consequently, the segmented patient surface is exploited to determine the points of interest for the autonomous ultrasound scan.

\subsection{System Calibration}
\label{sec:calibration}

\subsubsection{US Tool Configuration and Calibration}\label{sec:uscalib}
\begin{figure}[t]
	\centering
	\includegraphics[width=0.4\columnwidth]{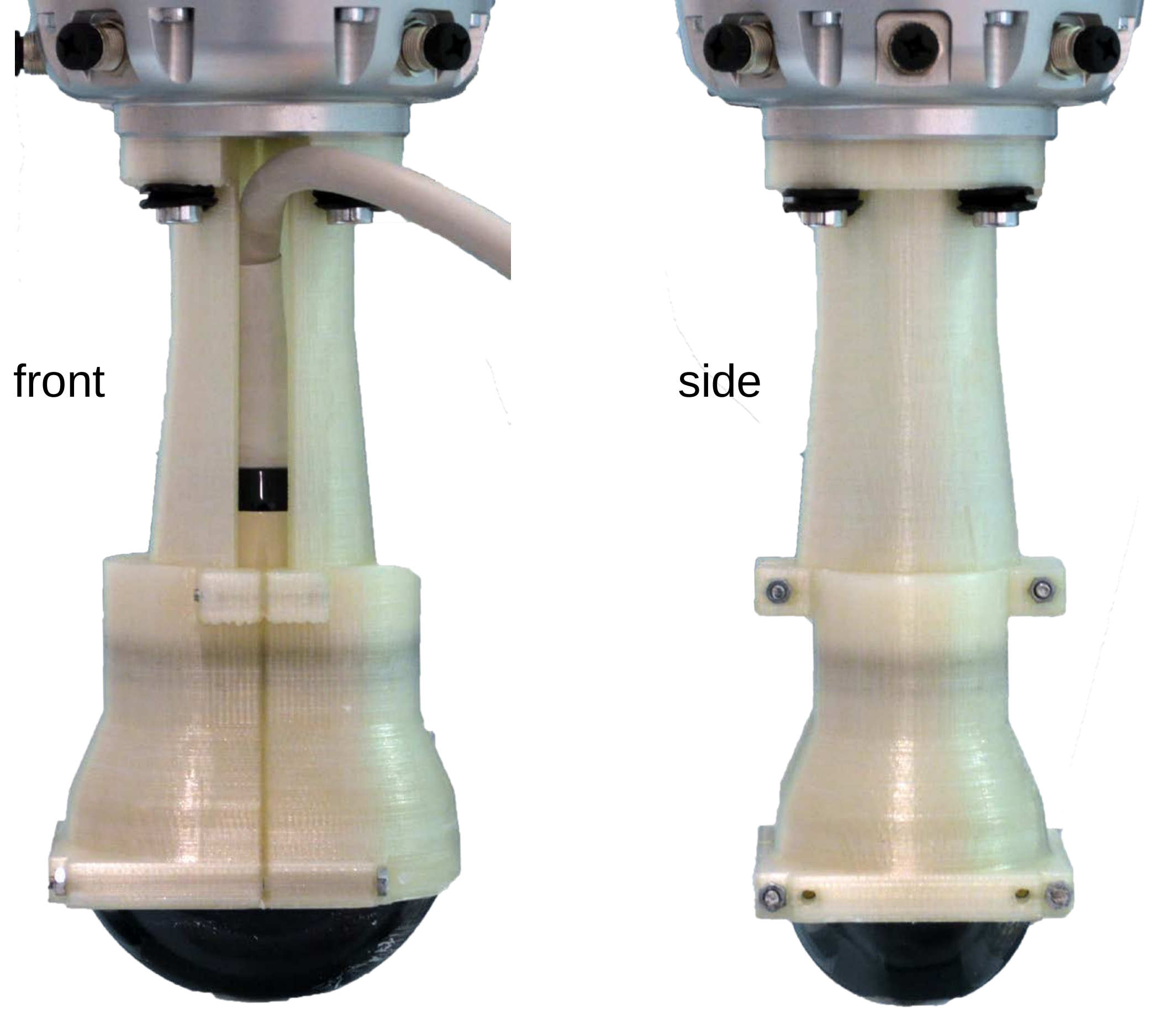}
	\caption{Ultrasound probe mount designed for lightweight arm end-effector.}
	\label{fig:probeMount}
\end{figure}

The transformation \superscript{T}\textbf{T}\subscript{E} from the robot end-effector to the transducer tool tip 
can be obtained directly from the CAD models used for manufacturing of a customized mount for the rigid attachment of the US probe to the robot end-effector. 
The probe mount utilized for the robotic system is shown in Fig. \ref{fig:probeMount} and is tailored to the ultrasound transducer for optimal and rigid attachment.
To perform ultrasound acquisitions, a second spatial transformation \superscript{US}\textbf{T}\subscript{T}, pointing from the probe tool tip to the US image origin, is defined by the image scaling (transformation from px to mm) and size with respect to the US probe apex
\begin{equation}
\superscript{T}T\subscript{US} = \left[
\begin{array}{cccc}
	s_x &    0 & 0 & s_x t_x \\
	0    & s_y & 0 & s_y t_y \\
	0    &    0 & 1 & 0   \\
	0    &    0 & 0 & 1   \\
\end{array}\right],
\end{equation}
where $s_x,s_y$ determine the resolution of the ultrasound image, and $t_x,t_y$ the translation from the apex, defined by the center of curvature, to the image origin.
If an additional refinement of the transformation from the ultrasound image to the transducer apex is required, a standard ultrasound to tracking calibration technique can be applied~\cite{mercier2005review}.

As the used lightweight robot provides force sensors in each joint, an additional calibration with respect to the weight and center of mass of the tool is necessary to allow the force controlled motion.
For an accurate load calibration, a proprietary calibration algorithm provided with the robot is used to determine these values in 3D. \\


\subsubsection{Camera-to-World calibration}\label{sec:robotworldcalib}
This calibration step allows for the control of the robot - representing the world reference frame - as observed in camera coordinates.
In the case of RGB-D cameras, this transformation relates from the camera to the robot arm, and can be computed either by using the 2D RGB images or the 3D information obtained by the depth sensor. 
An accurate calibration can be achieved with both techniques, although the 3D case requires more effort, user interaction and processing time \cite{kahn2014hand}.
Alternately, in \cite{hennersperger2016a}, an additional optical tracking system is used to achieve a good calibration between robot and camera. 
However, by leveraging the 2D RGB images, it is possible to compute the transformation \superscript{C}\textbf{T}\subscript{W} between the robot base and the RGB-D camera by using a standard hand-eye calibration method~\cite{tsai1988real} in its \lq eye-on-base\rq~ variant, without requiring an additional optical tracking system.
To do so, an Augmented Reality (AR) marker is placed on the robot flange and the transformation between the marker and the camera estimated using the ArUco~\cite{garrido2014automatic} tracking library.
Several tracked poses combining random (non co-linear) movements around several rotation axis are recorded and used as correspondence pairs for calibration. 
The robot forward kinematics and the presented transformation \superscript{T}\textbf{T}\subscript{E} complete the chain of transformations from the camera reference frame to the US transducer. 
It should be noted that by using this calibration, the exchange of the US transducer with another one does not require a full camera recalibration, but only recomputing the last transformation of the chain between the end-effector and the US probe \superscript{T}\textbf{T}\subscript{E}.
Furthermore, the AR marker-based calibration is used for the camera-to-world calibration only, allowing for a fully marker-free data acquisiton and update.
The transformation between the RGB and the depth sensors of the Kinect can be obtained as proposed by \cite{herrera2012joint}.
For the hardware used, this relates to a translation of $2.5$~cm along the transverse axis of the camera.
\\

\subsubsection{Patient-to-World Calibration}\label{sec:patientworldcalib}
With respect to a fully autonomous acquisition, it is necessary to transfer trajectories planned in a tomographic image to the actual position and orientation of a patient lying on the bed.
To do so, acquired 3D RGB-D information of the patient (Sec. \ref{sec:rgbdScanner}) provides surface information, which can be extracted simultaneously from the tomographic images.
These data can then be used directly for aligning both surfaces and consequently determining their spatial transformation.
In the following section, we will describe the necessary steps to enable such a calibration in detail.

\paragraph{Surface Extraction from MRI Data}
To extract the patient (or phantom) skin surface from the MRI images, we employ an approach which does not require complex segmentations, as the extracted and matched surfaces will be refined throughout the acquisition process using image-based registration. 
Therefore, tomographic data  (in this work T2-weighted MRI) is thresholded first, neglecting all values $I<\tau$. Throughout all our experiments, $\tau=100$ provided a sufficient discrimination between the patient and the background (acquisition table and room around patient).
As the resulting mask contains holes and is partially irregular, a morphological closing is performed on the thresholded image containing of subsequent dilation and erosion.
The surface can then be retrieved from the image data~\cite{gonzalez2004digital}, where only the connected surface component covering the highest ratio of total pixels is kept for further processing
\begin{equation}
	\Omega_B = \underset{i}{arg \: max}\sum \Omega_i, \quad \Omega_i = \{x_1, x_2, ... , x_m\},
\end{equation}
with $x_m$ being the surface positions of the points contained in the component $\Omega_i$ in 3D space.

\paragraph{Spatial Change Detection in RGB-D Data} 
To separate the patient from the background and other objects within the view of the camera, a spatial change detection is performed. 
Octrees are often used in computer graphics to partition a volume by recursively dividing cubes~\cite{elseberg2013one}. 
First, a background point cloud is recorded and added to an octree. 
After positioning the patient on the bed, a second point cloud is retrieved and added to another octree. 
The differences in the tree structure allow the efficient determination of newly added points, which represent the background-subtracted object, in our case the patient. 
To increase robustness and compensate for noise, a minimum of $n$ pixels (experimentally set to 2) must be contained in each tree node. 


\paragraph{Surface Matching}
The alignment of surfaces can either be achieved using Iterative Closest Points (ICP), a non-linear derivative of ICP, or by directly aligning features. 
As the patient surface shape deviates strongly from the shape of the background (e.g. table or ultrasound scanner) a feature alignment process is applicable. 
On the foundation of the calibration refinement using intensity-based registration (Sec. \ref{sec:patientupdate}), our framework automatically accounts for local deformations and inaccuracies.
Consequently, we employ ICP as surface matching method, as it provides a robust and especially highly efficient global alignment, which is then used to initialize the intensity-based registration.

Result of the surface-matching between extracted MRI and RGB-D information will be a rigid transformation from MRI-space to RGB-D camera space
\begin{equation}
P_{i}^{C} = \tensor[^{C}]{\textbf{T}}{_{MRI}} P_{i}^{MRI} = (\tensor[^{MRI}]{\textbf{T}}{_{C}})^{-1} P_{i}^{MRI},
\end{equation}
with $P_i^{MRI}, P_i^{C}$ being the surface points in MRI and camera-space respectively.

\subsection{Autonomous US Acquisition}\label{sec:acquisition}
Following the workflow as shown in Fig. \ref{workflow}, this section will describe all steps carried out to perform one or multiple autonomous acquisitions.

\subsubsection{Planning of Robotic Acquisition}\label{robot}
Based on the alignment of both camera and patient within the world reference (Sec. \ref{sec:calibration}), the next step is to transfer the previously planned image trajectory to a robotic control trajectory which can be executed by the robotic arm. To allow accurate and safe trajectories, a proper force control strategy of the robotic manipulator, and the planning of the acquisition are necessary.

 \paragraph{Stiffness and Force Control} 
The manipulator is mainly operated using a stiffness controller, which represents a Cartesian virtual spring between the desired (setpoint) position $x_{set}$ and current (measured) position $x_{msr}$. The control law for this strategy is defined by the transposed Jacobian matrix $J^T$
\begin{equation}
	\tau_{Cmd} = J^{T}(k_{c}(x_{set} - x_{msr})+D(d_c))+f_{dyn}(q,\dot{q},\ddot{q}),
	\label{eq:stiffness}
\end{equation}
where $k_c$ is the Cartesian stiffness of the virtual spring $k_c(x_{set} - x_{msr})$. The damping term $D(d_c)$ is dependent on the normalized damping value, while the dynamic model of the manipulator $f_{dyn}(q,\dot{q},\ddot{q})$ is predefined by the manufacturer~\cite{schreiber2010fast}.
The resulting joint torque is computed by a Cartesian law using $J^{T}$.

In order to allow for compliant force applied in the US probe direction, the force control is modified such that the force term $D(d_c)$ is set to a high damping in the respective probe direction, allowing for a compliant movement.
Thus the high accuracy torque sensors and real-time forward kinematics computation of the system are exploited in this view in order to provide an acoustic force coupling without applying excessive forces ($>25$~N) on the skin. 
The stiffness controller allows to achieve an indirect force control \cite{salisbury1980active}, considering the relationship between the deviation of the end-effector position and orientation from the planned motion as well as the contact force and moment.
The online parametrization of the virtual spring values makes it possible to maintain the contact force constant.
The Cartesian stiffness along the direction of the US probe is set in the range [125-500]~N/m according to the anatomy of the patient, and the force desired is parametrized as 5~N. The stiffness and forces in the other directions are parametrized to 2000~N/m and 0~N accordingly.
If an excessive force $> 25$~N occurs, the robot's internal subroutine will automatically stop the acquisition process.
This feedback loop enables the compliant motion constrained by the patient or other objects.

\paragraph{Planning of Acquisition Path} Based on the selected points (see Section \ref{sec:mri}) and corresponding surface normals provided by the RGB-D data, the acquisition path is planned. 
The selected start and end points of the trajectory $P_s$,$P_e$ define the direction vector of the trajectory (see Eq. \eqref{eq:trajectoryDir}), which is used to define equidistant sampling points $Q^1, Q^2, \ldots, Q^i$ along the line with a distance of $2$~cm.
For each sampling point, the closest surface point $Q^i \rightarrow P^k$ is retrieved by a nearest-neighbor search.
Along with the corresponding surface normal directions $n_k$, the points can then be used directly as target tool poses for robotic movements. 
It should be noted that the force controller implicitly commands the robot to adapt the tool position until the surface is reached.
As the robot approaches the next trajectory point $P^{k+1}$, the direction of the transducer is stepwise changed to the subsequent surface normal position defined by $P^{k},P^{k+1}$ by the robot control.

\subsubsection{US Acquisition and 3D Reconstruction}\label{sec:us}

The live streams of robot tracking data and ultrasound images are transmitted via Ethernet, therefore the offset between tracking and US data is small compared to the framework overhead. 
In order to enable 3D registration with the diagnostic image, a volume compounding is performed using a backward normalized-convolution approach following~\cite{hennersperger2014vascular}, yielding regular-spaced 3D volume data from the arbitrarily sampled ultrasound frames in 3D space.

\subsubsection{US-to-MRI Alignment}\label{sec:usmralignment}

Making use of all previously estimated transformations from camera to MRI $\tensor[^{MRI}]{\textbf{T}}{_{C}}$, world to camera $\tensor[^{C}]{\textbf{T}}{_W}$ and ultrasound to tool $\tensor[^{T}]{\textbf{T}}{_{US}}$, we can transform both ultrasound and MRI data into the world space using the respective transformations from both image spaces to the world reference frame
\begin{align}
\tensor[^{W}]{\textbf{T}}{_{MRI}} &= {(\tensor[^{MRI}]{\textbf{T}}{_{W}})}^{-1} =   
(\tensor[^{MRI}]{\textbf{T}}{_{C}} \cdot \tensor[^{C}]{\textbf{T}}{_W})^{-1} \\
\tensor[^{W}]{\textbf{T}}{_{US}} &= \tensor[^{W}]{\textbf{T}}{_E} \cdot \tensor[^{E}]{\textbf{T}}{_T} \cdot \tensor[^{T}]{\textbf{T}}{_{US}}.
\label{equ:transChain}
\end{align}
Both image datasets will be roughly aligned after transformation into the world coordinate frame, relying on prior ultrasound and patient to world calibrations.

\subsubsection{MRI-to-US Registration}\label{sec:mrusregistration}
 Supposing a rough overlap of MRI and US images as described above, an intensity-based MRI/US registration can directly be initialized in order to obtain an updated transformation between ultrasound and tomographic images.

The LC\superscript{2} similarity method allows for the registration of MRI and US images by correlating the MRI intensities and gradient magnitudes to the US intensity values. 
High robustness, wide convergence range and the application for rigid, affine and deformable registration have been shown in~\cite{fuerst2014automatic}. 
This approach is utilized in a two-step process.
Based on the calibration and transformation chain, we can bring recorded ultrasound volume data directly into the MRI reference frame (or both into the world reference frame)
\begin{equation}
\tensor[^{MRI}]{\textbf{T}}{_{US}} = \tensor[^{MRI}]{\textbf{T}}{_{W}} \cdot \tensor[^{W}]{\textbf{T}}{_{US}} = (\tensor[^{W}]{\textbf{T}}{_{MRI}})^{-1} \cdot \tensor[^{W}]{\textbf{T}}{_{US}}.
\end{equation}
In a second step, LC\superscript{2} is used to determine the rigid transformation component, aligning the transformed US and MRI images in world space in reference to the actual patient position
\begin{equation}
 \tensor[^{P}]{\textbf{T}}{_{W}} = \tensor[^{P}]{\textbf{T}}{_{MRI}} \tensor[^{MRI}]{\textbf{T}}{_{W}},
 \label{equ:usMriReg}
\end{equation}
with $\tensor[^{P}]{\textbf{T}}{_{MRI}}$ being the updated transformation to compensate for patient movement, deformation, as well as tracking and detection inaccuracies of $\tensor[^{C}]{\textbf{T}}{_{MRI}}$ with respect to the world reference frame, \cf ~Eq. \eqref{equ:transChain}.
Optionally, a second step consisting of an affine and deformable registration using the same similarity measure can be performed, resulting in a precise and direct correspondence between voxels in the compounded 3D-US and reconstructed MRI volumes, which finally closes the loop of the autonomous ultrasound acquisition system and enabling MRI guided interventions. 
We evaluate both rigid and affine registrations; the decision whether both are required relies on the specific application in mind.
For instance, if bony structures are to be scanned rather than soft tissue, a rigid alignment would be sufficient based on our experience.

\subsubsection{Update of Patient Calibration}\label{sec:patientupdate}
Based on the estimated transformations from the (robotically) acquired 3D ultrasound data and the MRI aligned to the world, the transformation from tomographic imaging to ultrasound space can be refined, in order to enable a more precise initialization of subsequent acquisitions.
This becomes especially interesting as such acquisitions can be performed for an on-line refinement of the whole system calibration through an image-based feedback and update.
In an interventional setup, an initial acquisition is performed at the beginning to optimize the robot- and patient-to-world calibrations.
Subsequent planned trajectories can then be performed automatically based on an updated calibration. 
Making use of the estimated rigid transformation ${\tensor[^{P}]{\textbf{T}}{_{MRI}}}$ aligning the US and MR volumes in world space as described above, the patient-to-world calibration can be updated accordingly to

\begin{equation}
\tensor[^{P}]{\textbf{T}}{_{US}} = \tensor[^{P}]{\textbf{T}}{_{MRI}} \cdot \tensor[^{MRI}]{\textbf{T}}{_{C}} \cdot \tensor[^{C}]{\textbf{T}}{_{W}} \cdot \tensor[^{W}]{\textbf{T}}{_{US}}.
\end{equation}
On the one hand, this reduces processing time, as no rigid alignment is required for those US volumes.
On the other hand, the comparability of subsequent records is fully maintained by this approach, as images are provided for the identical anatomy and planned trajectory.
For an exemplary application of US-guided liver biopsy, several acquisitions of the liver could be conducted throughout the procedure, enabling a reliable 3D image-guidance during needle insertion. 
%
%
%
%
\section{Experiments and Results}\label{sec:experiments}
%
%

\subsection{Robot/Camera Calibration and Robot Control}

\begin{figure}[]
	\centering
	\includegraphics[width=\columnwidth]{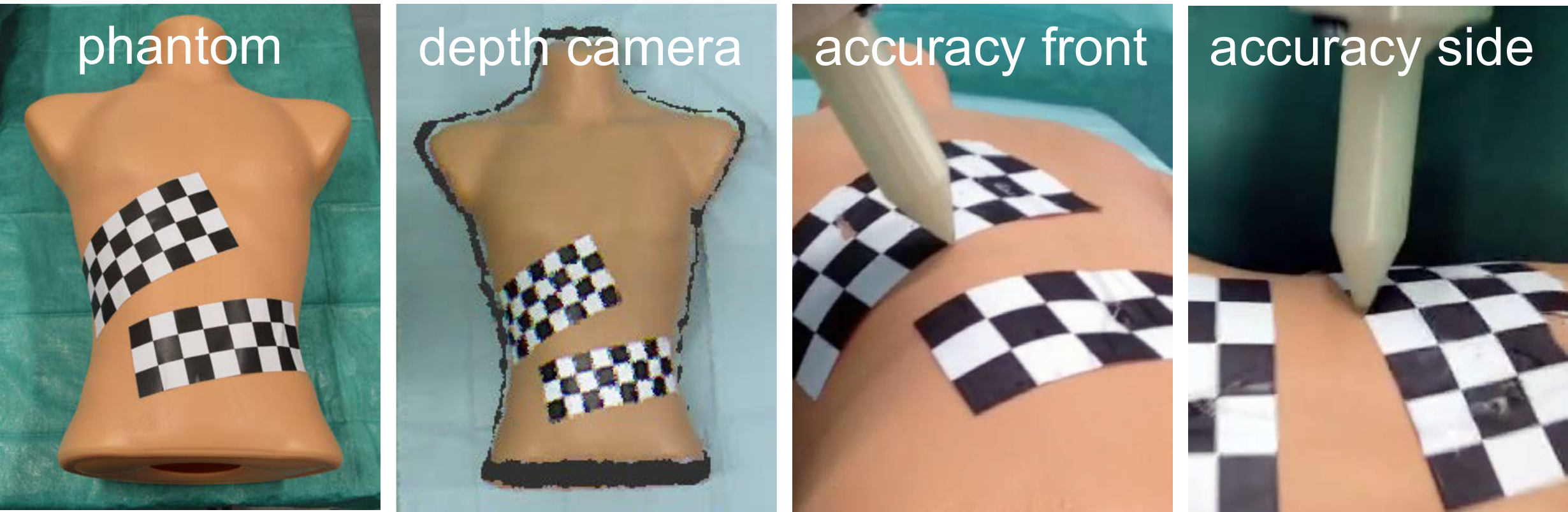}
	\caption{From the colored point cloud (depth camera), individual checkerboard corner points on the uneven surface are manually selected as target points. The points can be targeted with an accuracy of $2.46$~mm in x-y-direction, and $6.43$~mm in z-direction.}
	\label{fig:resultsSurface}
\end{figure}

Utilizing the depth images augmented with color information, we evaluate the accuracy of the tool and camera-to-world calibrations by equipping the robot manipulator with a rigid acuminate tool and moving the tool tip onto selected points on a realistic rigid surface. 
In this view, after tool configuration, the camera-to-robot calibration is performed to obtain the required transformations as described in Sec. \ref{sec:robotworldcalib}, where 13 poses are used in total for the calibration.
Next, the RGB-D data of an upper torso phantom surface is recorded and the surface normals computed. 
The planned acquisition path is defined by manually selecting multiple intersection points on a printed checkerboard attached to the  phantom's curved surface, see Fig. \ref{fig:resultsSurface}.
The robot is commanded in position control mode to move onto the individual points and the distances to the actual intersection points are measured manually with a caliper.
The experiment is performed twice by selecting 10 points in each run and recalibrating the whole system in between the two sessions.
For the first run, the results yielded an average accuracy $|x_{set} - x_{msr}|$ of $2.42\pm 1.00$~mm on the x-y plane and $7.20\pm 3.30$~mm along the z-axis (camera's depth axis). 
For the second one, the accuracy was estimated as $2.5\pm 0.90$~mm (x-y plane) and $5.64\pm 4.04$~mm (z-axis). 
Overall, accuracies were $2.47 \pm 0.96$~mm (x-y) and $6.43 \pm 3.68$~mm (z).
During all experiments, the camera was placed at around $1.5$~m distance to the phantom.
This shows that the calibration yields reproducible results, while the inaccuracies are dependent on the RGB-D camera information.
It should be noted that these values are in line with the reported average spatial resolution of the camera, being $3.0$~mm (x-y plane) and $10$~mm (depth axis) for a camera at $2.0$~m from its target~\cite{andersen2015kinect}.
It is also important to notice that while the $x-y$ accuracy directly affects the robot poses, the $z$ errors are effectively compensated by the desired-force control of the robot. In this view, the resulting system layout can compensate for the inaccuracies of the RGB-D camera.
Based on these accuracies, the calibration and overall system accuracy is anticipated to be sufficient for the initialization of an image-based registration, and thus also for the full system.

\subsection{MRI/US Image Acquisition and Registration}

To allow for an evaluation of the overall system, we first use a triple-modality 3D abdominal phantom (Model 057A, Cirs Inc., Norfolk, Virginia, USA), which provides anatomically correct objects such as the liver, cysts, ribs and vertebrae.
The phantom is targeted at MRI, CT, and ultrasound imaging and provides realistic image properties for all modalities.
We then perform similar automatic robotic US acquisitions on two healthy volunteers, for whom an MRI scan was performed prior to this work.

\begin{figure}
	\centering
		\includegraphics[width=0.49\columnwidth]{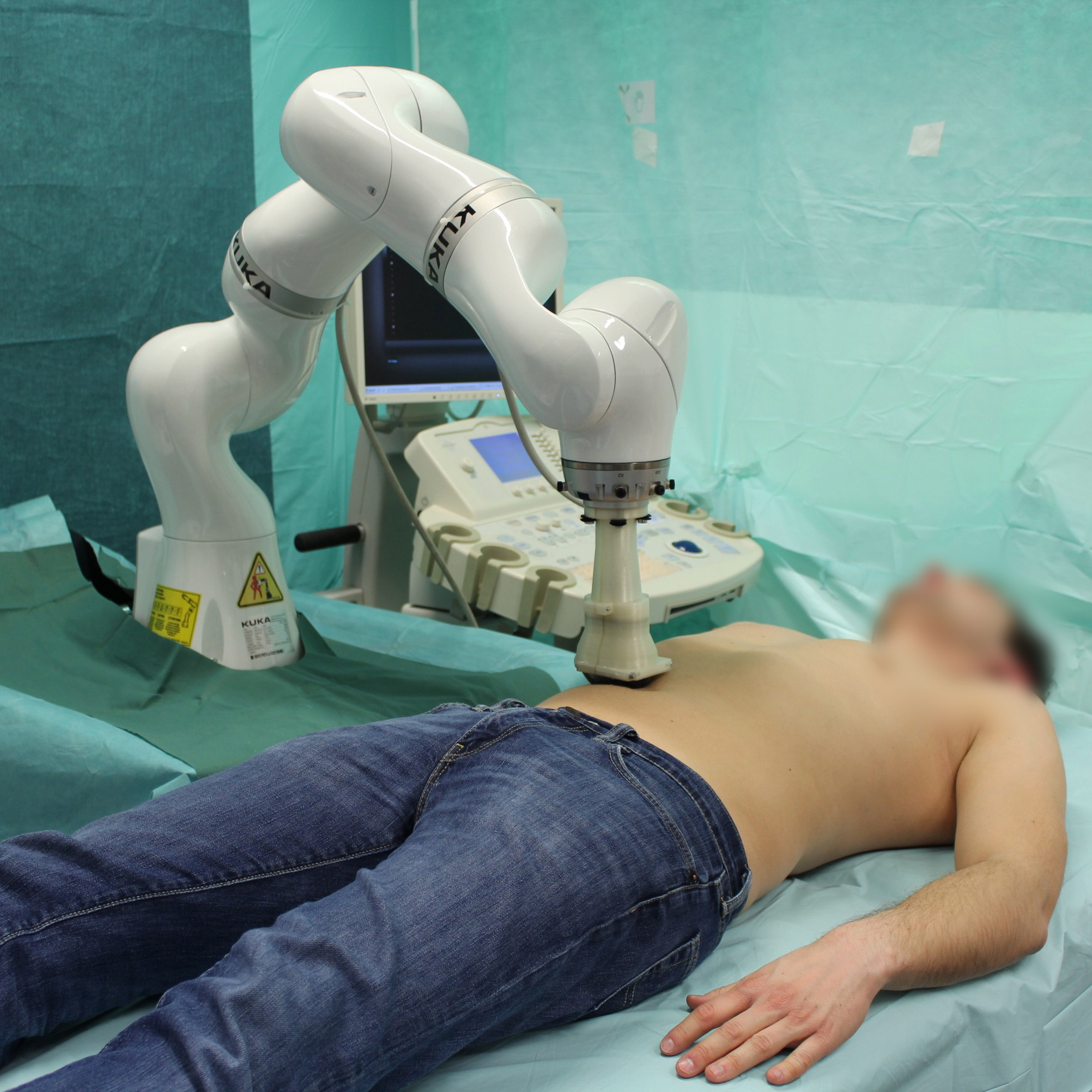}
		\includegraphics[width=0.49\columnwidth]{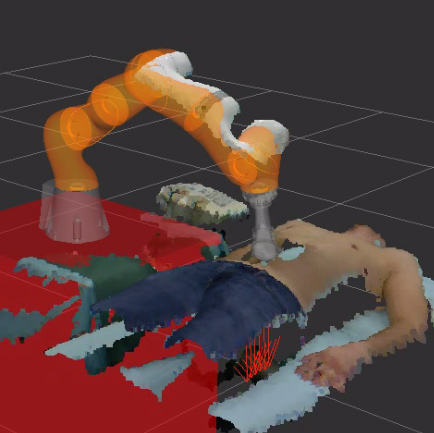}
	\caption{Scanning setup showing the real system setup scanning a volunteer (left) as well as the visualization of all involved components accordingly with the point cloud given by the RGB-D camera.}
	\label{fig:scanningSetup}
\end{figure}

Following the workflow presented, the MRI images are transformed into the world reference frame by matching the 3D scanned surface with the MRI surface (see Sec. \ref{sec:patientworldcalib}). 
Based on this global registration, trajectories are planned in the MRI by selecting start- and endpoint.
Figure \ref{fig:scanningSetup} shows the actual scanning setup for a healthy volunteer, where the robot arm performs an automatic acquisition based on a planned trajectory.
Given a perfect system accuracy, i.e. a system with perfect calibration and without any tracking or imaging inaccuracies, an automatic acquisition of a 3D ultrasound volume by the robotic system would yield a perfect overlap between reconstructed 3D US data and tomographic imaging.
In reality, however, the image alignment will be affected, which is why an intensity-based registration is then performed in order to detect the offset and correct for system inaccuracies. 
The alignment by intensity-based registration is first performed first rigidly (accounting for tracking, calibration, patient registration, and movement), followed by an affine improvement (accounting for image scaling as well as patient and tissue movement/deformation). 
Due to the fixed direction of the trajectory and a constant skin-pressure, an affine registration is sufficient for this application. 
The resulting rigid transformation directly indicates the accuracy of the overall 3D scanning system based on the alignment of planned and real acquisition (distance between desired and actual US). 
In a second step, we then use the rigid transformation part of the registration to refine the overall system and robot calibration according to Eq. \eqref{equ:usMriReg}.
The experiment is repeated with a second planned trajectory, followed by an evaluation of rigid and affine registrations between the calibration-updated scans and the tomographic data.
The hypothesis is that by using the rigid part of the intensity-based registration, we can align the pre-interventional MRI with the interventional US data. 
This closes the loop between world calibration and robotic-system, where the system calibration can be refined online.
In order to also evaluate the errors of the system in different situations, we repeat the overall experiment two times for the phantom, and between each experiment the phantom is moved on the examination bed.
For the scans on volunteers, we repeat the overall experiment three times per person, where the person stands up and lies down between each run, evaluating also potential variabilities based on the patient-to-world calibration.
Robotic acquisitions are taken at a similar speed as freehand acuqisitions, such that a normal acquisition requires $<30$~s.
Ultrasound scans are planned in the the abdominal region and contain organs (e.g. liver, kidney) and vessels.
Acquisitions may also contain ribs, where the robot follows the natural curvature of the body surface without exceeding the maximal force applied. 
Ultrasound parameters are set manually, although modern US systems provide automatic adaption allowing for a direct optimization of settings. 
Coupling gel is also applied manually, and the volunteers must hold their breath during the scan.	
Safety of acquisitions is guaranteed by the collision detection provided by internal subsystems incorporated in the KUKA lightweight system, which would stop actions automatically if the force desired is exceeded.
The patient-to-world calibration is performed once before each acquisition, but could be also updated online in future.
The global alignment between the patient and MRI surface requires less than 10s, and the intensity-based refinement of the registration requires less than 30s to complete.

\begin{figure}[t]
	\centering
		\includegraphics[width=0.5\columnwidth]{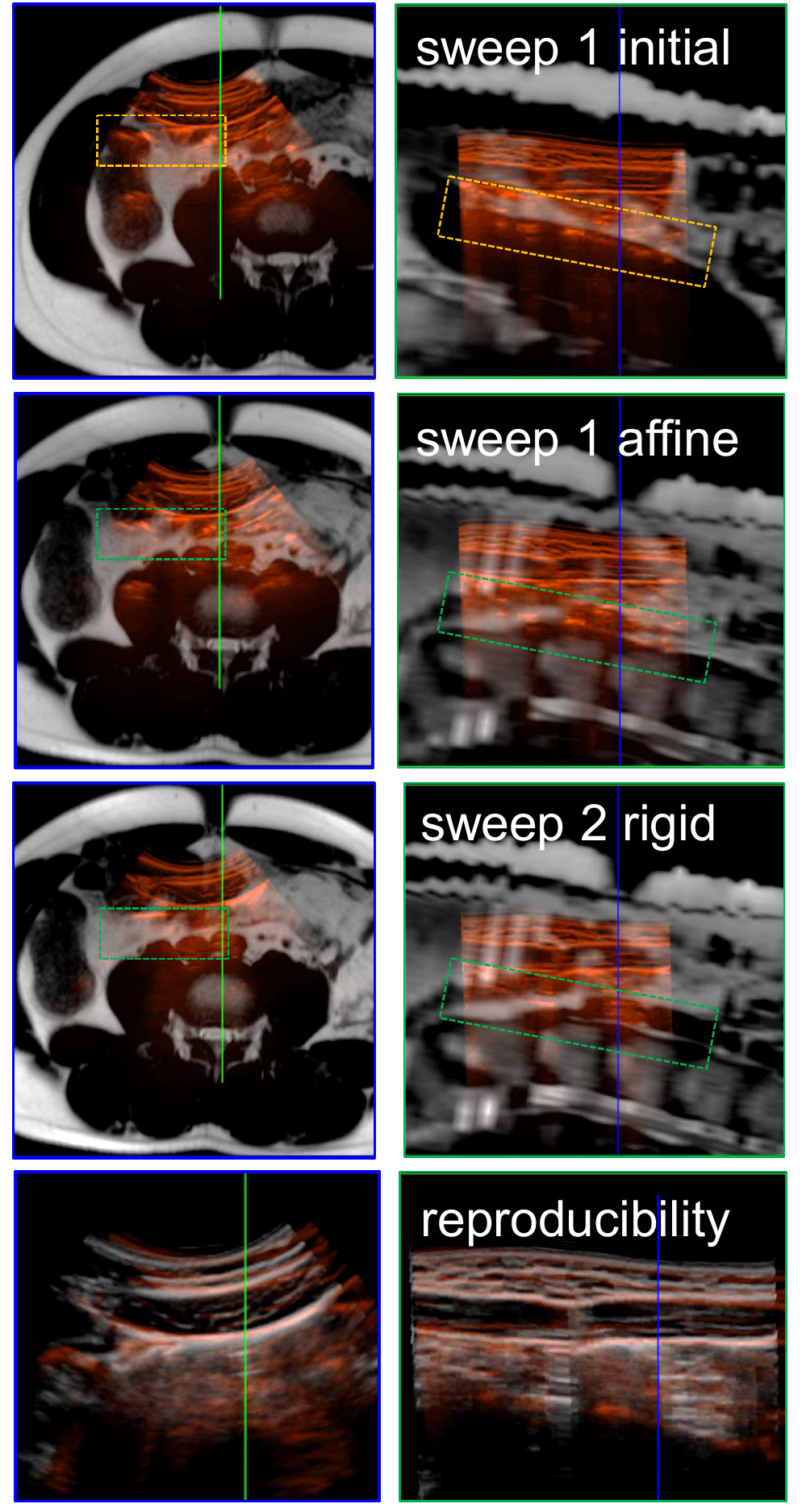}
	\caption{Left column: axial plane; right column: saggital plane. The initial calibration shows partial misalignment (orange box), while the registrations for sweep \#1 and \#2 account for this offset. An overlay of both scans also shows the high reproducibility of the sweeps.}
	\label{fig:qualitativeResults}
\end{figure}

For all acquisitions of the phantom and humans combined, the average translation between the initial US-volume and MRI is $4.47 \pm 2.15$~mm for iteration one (without calibration-update) compared to $0.97 \pm 0.76$~mm in the second run after the overall patient calibration is updated.
Similarly, the rotational part clearly improves after the update of the calibration ($4.50 \pm 2.24^\circ$ before update, versus $1.08 \pm 0.76^\circ$ after), where the rotational error is determined as the Euclidean norm of the three rotation angles.
For an additional affine registration after the initial rigid registration, the average translation results in $1.02 \pm 0.77$~mm, which shows that the initial registration successfully accounts for potential tracking and point cloud inaccuracies (rotation errors are $1.43 \pm 1.42^\circ$).
Resulting ultrasound datasets and registrations are depicted in Fig. \ref{fig:qualitativeResults} for a volunteer acquisition, and all results of the different experiments are listed in Table 1.
%
%
\begin{equation*}
\centering
\begin{array}{|c||c|c|}
\hline
& \multicolumn{2}{c|}{\textrm{Translation [mm]}} \\
&  \textrm{Phantom} & \textrm{Human}  \\
\hline
\textrm{Rigid scan \#1} & 4.69\pm1.35 & 4.39\pm2.47 \\
\textrm{Rigid scan \#2} & 1.18\pm0.52 & 0.89\pm0.84\\
\textrm{Affine scan \#1} & 1.30\pm1.06& 0.92\pm0.75\\
\hline
& \multicolumn{2}{c|}{\textrm{Rotation [}^\circ\textrm{]}} \\
\hline
\textrm{Rigid scan \#1} & 3.66\pm1.99 & 4.79\pm2.42 \\
\textrm{Rigid scan \#2}  & 0.40\pm0.32 & 1.30\pm0.74 \\
\textrm{Affine scan \#1} & 0.46\pm0.19 & 1.37\pm1.60\\
\hline
\multicolumn{3}{c}{\textrm{Table 1. Quantitative results based on registration}}
\end{array}
\label{tab:quantitativeResults}
\end{equation*}
Thereby, initial, rigid and affine refer to the initial US scan, the registration based on the calibration update, and the affine registration after rigid registration, respectively.

%
%
%
\section{Discussion}\label{sec:discussion}
%
%
Based on the millimeter accuracy achieved on repeated acquisitions with an updated calibration, our results suggest the feasibility of the overall approach as well as a potential path for a clinical integration of the presented system. 
The initial system-calibration using RGB-D information showed a maximum error below $1$~cm, which ensures that the acquired 3D-US datasets will lie within the capture range of the intensity-based registration~\cite{fuerst2014automatic}.
Therefore, the system should be able to deal with challenging clinical settings, where a higher deviation might lead to local minima during the registration optimization.
As the position and orientation of the acquired ultrasound images are aligned with the diagnostic image, US data can be correlated directly to the underlying tomographic image data, aiding the medical expert in terms of ultrasound visualization.
The results of the intensity-based registration (Table 1) show the overall translation/rotation of the US volume with respect to the MRI data, where the affine registration of the second scan appeared to perform worse than the rigid registration after calibration-refinement.  
While the quantity of data is too little to allow for a statistical meaningful judgement, the local deformation of tissue by the robotically held US probe is the primary reason for this discrepancy in our experience.
In view of a specific clinical application in mind, it is thus important to analyze and adapt the utilized registration tools further, especially with respect to a deformable registration of soft tissue.

The integration of ROS and Sunrise.OS as used in this work allows for a full control of the KUKA APIs directly from components being available within ROS. 
Furthermore, a direct access to the robot state is provided by the KUKA APIs, including native features such as self collision avoidance. 
In this regard, the utilized approach allows for the best combination of both worlds.
This facilitates the rapid development of new approaches, but also ensures safety of the robot environment, i.e. by collision detection and emergency halt features.

In view of a clinical integration, it should be noted that our results provide only a first feasibility evaluation, where future work clearly needs to focus not only on healthy volunteers but also diseased and pathological anatomy.
Methods such as patient surface extraction and dynamic registration will require further adaptions, aiming at a clinical integration in the future.
Our experiments also showed that the currently used examination bed is not optimal for the robot working space, as the end-effector is moving almost at the base of the robot.
With respect to a clinical application, we thus suggest a height-adjustable bed, which could be directly integrated with the imaging system, such that the optimal height would be adjusted based on the planned acquisition.
Besides that, acoustic coupling between the US probe and patient surface needs to be explored, facilitating US scans with either automatic gel application, or the exploration of other alternatives.

To this end, our future work will also focus on the possible online-optimization of the trajectory based on the initial planning, as we partially experienced suboptimal US image quality for the selected trajectories in this work.
In this view, also an automatic change of the applied coupling force with respect to the target surface (e.g. fat vs. muscle) is analyzed in ongoing efforts. 
%
%
\section{Conclusion}\label{sec:conclusion}
%
%
We have presented a path to an autonomous robotic ultrasound system in order to enable imaging and support during interventions. 
The set of methods presented shows the basic feasibility of the automatic scanning framework, allowing fast and efficient robotic 3D ultrasound acquisitions based on pre-interventional planned trajectories.
On the foundation of an integrated system consisting of a baseline tomographic image with 3D RGB-D information, we automatically register patient data and perform automatic robotic acquisition.
We introduced a closed-loop calibration update based on image-based registration to facilitate the acquisition of reproducible 3D US data.
Our results show that the overall accuracy of the system is sufficient for clinical applications. Despite challenges which need to be overcome before such systems could be used in daily routine, this work will hopefully facilitate the clinical acceptance of automatic and autonomous robotic ultrasound scanning systems in the future.

\section*{Acknowledgment}
We thank the department of nuclear medicine at Klinikum rechts der Isar for the MRI acquisitions, as well as ImFusion GmbH for the provided registration software. 
This work was partially supported by the Bayerische Forschungsstiftung project RoBildOR.


\ifCLASSOPTIONcaptionsoff
  \newpage
\fi

\bibliographystyle{IEEEtran}
\bibliography{IEEEabrv,literature/literature}

\end{document}